\def\year{2022}\relax
\documentclass[letterpaper]{article} 
\usepackage{aaai22}  
\usepackage{times}  
\usepackage{helvet}  
\usepackage{courier}  
\usepackage[hyphens]{url}  
\usepackage{graphicx} 
\usepackage{natbib}  
\usepackage{caption} 
\DeclareCaptionStyle{ruled}{labelfont=normalfont,labelsep=colon,strut=off} 
\usepackage[ruled,vlined]{algorithm2e}

\usepackage{xcolor}
\usepackage{hyperref}
\hypersetup{%
  colorlinks=true,
  urlcolor=blue,%
  linkcolor=blue,%
  citecolor=blue,%
  linkbordercolor=blue,
  pdfborderstyle={/S/U/W 1}
}

\title{Fairness without Imputation: \\A Decision Tree Approach for Fair Prediction with Missing Values}

\author{
    Haewon Jeong, Hao Wang, Flavio P. Calmon
}
\affiliations{
    Harvard University\\
    \{haewon,flavio\}@seas.harvard.edu, hao\_wang@g.harvard.edu
}

\usepackage{booktabs}    
\usepackage{multirow}

\newcommand{\sccell}[2]{\setlength{\tabcolsep}{0pt}\text{\begin{tabular}{#1}#2 \end{tabular}}}

\usepackage{graphicx}
\usepackage{subcaption}
\usepackage{mathrsfs}
\usepackage{amsfonts}

\usepackage{amsthm}
\usepackage{amssymb}
\usepackage[cmex10]{amsmath}
\usepackage{bm}

\usepackage{mdframed}
\usepackage{dsfont}

\def\mbP{\mathbf{P}}

\def\mbc{\mathbf{c}}

\def\mbp{\mathbf{p}}
\def\mbq{\mathbf{q}}

\def\mbu{\mathbf{u}}

\def\mbw{\mathbf{w}}
\def\mbx{\mathbf{x}}

\def\mbz{\mathbf{z}}

\def\calD{\mathcal{D}}

\def\calL{\mathcal{L}}

\def\calT{\mathcal{T}}

\def\calV{\mathcal{V}}

\def\binset{\{ 0, 1 \}}

\newtheorem{thm}{Theorem}
\newtheorem{lem}{Lemma}

\theoremstyle{definition}

\newtheorem{defn}{Definition}

\newtheorem{rem}{Remark}

\DeclareMathOperator*{\argmin}{\arg\!\min}

\newcommand{\mean}{\mathbb{E}}
\newcommand{\ExpVal}[2]{\mean{}\left[ #2 \right]}
\newcommand{\EE}[1]{\ExpVal{}{#1}}
\newcommand{\var}[1]{\mbox{Var}\left[#1\right]}

\newcommand{\textfn}[1]{{\small\textit{#1}}}

\newcommand{\defined}{\triangleq}

\newcommand{\sto}{\mbox{\normalfont s.t.}}
\newcommand{\indicator}[1]{\mathbb{I}{#1}}

\newcommand{\TV}{\textnormal{D}_{\mathsf{TV}}}

\newcommand{\Reals}{\mathbb{R}}

\newcommand{\dif}{\textrm{d}}

\usepackage{float}
\floatstyle{ruled}
\newfloat{program}{thp}{lop}
\floatname{program}{Program}

\begin{document}

\maketitle

\begin{abstract}

We investigate the fairness concerns of training a machine learning model using data with missing values. Even though there are a number of fairness intervention methods in the literature, most of them require a complete training set as input. In practice, data can have missing values, and data missing patterns can depend on group attributes (e.g. gender or race). Simply applying off-the-shelf fair learning algorithms to an imputed dataset may lead to an unfair model. In this paper, we first theoretically analyze different sources of discrimination risks when training with an imputed dataset. Then, we propose an integrated approach based on decision trees that does not require a separate process of imputation and  learning. 
Instead, we train a tree with missing incorporated as attribute (MIA), which does not require explicit imputation, and we optimize a fairness-regularized objective function.
We demonstrate that our approach outperforms existing fairness intervention methods applied to an imputed dataset, through several experiments on real-world datasets.

\end{abstract}

\section{Introduction}

Datasets can contain missing values, i.e., unobserved variables that would be meaningful for analysis if observed~\citep{little2019statistical}. In many domains ranging from survey data, electronic health records, and recommender systems, data missingness is so common that it is the norm rather than the exception. This challenge has inspired significant research on  methods to handle missing values~\citep{little2019statistical,schafer2002missing,buuren2010mice,royston2004multiple,molenberghs2007missing}. 
Data missingness is usually modeled as a random process independent of non-missing features. However, when it comes to human-related data, data missingness often correlates with the subject's sociodemographic group attributes\footnote{We refer to attributes that identify groups of individuals (e.g., age, ethnicity, sex) as \emph{group attributes}.}. For example, in medical data, there can be more missing values in low-income patients as they are more likely to refuse to take costly medical tests. In survey data, questionnaires can be less friendly for a certain population, e.g., the fonts are too small for elderly participants, or the language is too difficult for non-native English speakers. 
We show a concrete example of different missing patterns in different sociodemographic groups in the high school longitudinal study (HSLS) dataset~\citep{ingels2011high}, which contains education-related surveys from students and parents\footnote{See Section C of the supplementary material for details}. In the parent survey, there are generally 6-7\% more missing values in under-represented minority (URM) students\footnote{In this paper we use the term URM to include Black, Hispanic, Native American and Pacific Islander.} compared to White and Asian students. On the questions related to secondary caregivers, URM students had 15\% more missing values, as there are disproportionately more children in a single-parent family in URM~\citep{census2019}.

In the machine learning (ML) pipeline, missing values are handled in the preprocessing step (either dropping missing entries or performing imputation) before training a model. When there exist disparate missing patterns across different groups, how missing values are treated can affect the final trained prediction model, not just its accuracy, but also its fairness. Despite the ubiquity of the problem, fairness issues in missing values have gained only limited attention~\citep{fernando2021missing,wang2021analyzing}. The burgeoning literature on fair ML algorithms either overlooks the existence of missing data or assumes that the training data are always complete \citep[see e.g.,][]{calmon2017optimized,hardt2016equality,zemel2013learning,zafar2019fairness,feldman2015certifying,menon2018cost}. 

In this work, we connect two well-studied, yet still mostly disjoint research areas---handling missing values and fair ML---and answer a crucial question: 
\begin{itemize}
    \item[] \emph{In order to train a fair model on data with missing values, is it sufficient to first impute the data then apply a fair learning algorithm?}
\end{itemize}
We address this question through both a theoretical analysis and an experimental study with real-world datasets.

In the first part of the paper, we examine the limitation of the disconnected process of imputing first and training next. We identify three different potential sources of discrimination. 
First, we show that depending on the missing patterns, the performance of imputation methods can be different per group. As a result, the predictive model trained on imputed data can inherit and propagate  biases that exists in the imputed data. Then, we show that even when we use an imputation method that has an unbiased performance at the training time, if different imputation is employed at the testing time, this can give rise to discriminatory performance of the trained model. Finally, we prove a fundamental information-theoretic result: \emph{there is no universally fair imputation method for different downstream learning tasks}.

To overcome the above-mentioned limitations, we propose an integrated approach, called \emph{Fair MIP Forest}, which learns a fair model without the need for explicit imputation. The Fair MIP Forest algorithm is a decision tree based approach that combines two different ideas: missing incorporated as attribute (MIA)~\citep{twala2008good} for handling missing values and mixed integer programming (MIP)~\citep{bertsimas2017optimal} formulation to optimize a fairness-regularized objective function. 
By marrying the two ideas, we are able to optimize fairness and accuracy in an end-to-end fashion, instead of finding an optimal solution for the imputation process and the training process separately. 
Finally, we propose using an ensemble of fair trees instead of training a single tree. This reduces overall training time and improves accuracy. We implement the Fair MIP Forest algorithm and test it on three real-world datasets, including the aforementioned HSLS dataset. The experimental results show that our approach performs favorably compared to existing fair learning algorithms trained on an imputed dataset in terms of fairness-accuracy trade-off.

\subsection{Related Works}

Methods that handle missing values have a long history in statistics \citep{little2019statistical,schafer2002missing,buuren2010mice,royston2004multiple,molenberghs2007missing,stekhoven2012missforest,tang2017random}.
The simplest way of handling missing values is dropping rows with missing entries (also known as complete case analysis). However, in the missing values literature, it is strongly advised to use all the available data as even with a small missing rate (e.g. 2-3\%), dropping can lead to suboptimal performance and unethical selection bias~\citep{newman2014missing}.
A more desirable way to deal with missing values is imputation~\citep{little2019statistical}, where missing entries are replaced with a new value. This includes inserting dummy values, mean imputation, or regression imputation (e.g. k-nearest neighbor (k-NN) regression)~\citep{donders2006gentle,zhang2016missing,bertsimas2017predictive}. 
Multiple imputation is a popular class of imputation methods~\cite{rubin2004multiple,wulff2017multiple,white2011multiple} that draws a set of possible values to fill in missing values, as opposed to single imputation that substitutes a missing entry with a single value.
While our theoretical analysis focuses on single imputation for its simplicity, our proposed Fair MIP Forest algorithm shares conceptual similarities with multiple imputation since we train multiple trees with different random mini batches, each of which treats missing values differently. 

Alternatively, some ML models do not require an explicit imputation step. Decision trees, for instance, have several ways to handle missing values directly such as surrogate splits~\citep{breiman2017classification},  block propagation~\citep{ke2017lightgbm}, and missing incorporated as attribute (MIA)~\citep{twala2008good}. In the second part of the paper, we focus on decision trees with MIA as it is empirically shown to outperform other missing values methods in decision trees~\citep{kapelner2015prediction,josse2019consistency}.

We study the problem of learning from incomplete data and its fairness implications. 
In this regard, our work is related with \citet{kallus2019assessing,fogliato2020fairness,mehrotra2021mitigating} which consider the case where the sensitive attributes or labels are missing or noisy (e.g. due to imputation). In contrast, we focus on the case where the input features are missing and may thus impact performance in downstream prediction tasks. 
The works that are most relevant to our work are \citet{fernando2021missing,wang2021analyzing} as they examine the intersection of general data missingness and fairness. While \citet{fernando2021missing} presents a comprehensive investigation on the relationship between fairness and missing values, their analyses are limited to observational and empirical studies on how different ways of handling missing values can affect fairness. \citet{wang2021analyzing} proposes reweighting scheme that assigns lower weight to data points with missing values by extending the preprocessing scheme given in~\citet{calmon2017optimized}. 
The question we address in this work is fundamentally different from these two works. We examine if we can rectify fairness issues that arise from missing values by applying existing fair learning algorithms after imputation. Unlike the previous works that are limited to empirical evaluations, we demonstrate the potential fairness risk of learning from imputed data from a theoretical perspective.
Our theoretical analysis is inspired by a line of works \citep[see e.g.,][]{chen2018my,feldman2015certifying,zhao2019inherent} which aim to quantify and explain algorithmic discrimination. 
Furthermore, our solution to learning a fair model from missing values is an in-processing approach that explicitly minimizes widely-adopted group fairness measures such as equalized odds or accuracy parity, in contrast to the preprocessing approach in \citet{wang2021analyzing}.

Our proposed Fair MIP Forest algorithm is inspired by \citet{aghaei2019learning} that introduced training a fair decision tree using mixed integer linear programming. Framing decision tree training as an integer program was first suggested in \citet{bertsimas2017optimal}, and adapted and improved in \citet{verwer2017learning,verwer2019learning}. Unlike conventional decision tree algorithms where splitting at each node is determined in a greedy manner, this formulation produces a tree that minimizes a specified objective function. We  follow the optimization formulation given in \citet{bertsimas2017optimal}, but we add MIA in the optimization to decide the optimal way to send missing entries at each branching node. To the best of our knowledge, our work is the first one to include missing value handling in the integer optimization framework. Furthermore, unlike \citet{aghaei2019learning} which only had statistical parity as a fairness regularizer, we implement four different fairness metrics (FPR/FNR/accuracy difference and equalized odds) in the integer program. Finally, we go one step further from training a single tree, and propose using an ensemble of weakly-optimized trees. \citet{raff2018fair} also studies a fair ensemble of decision trees. However, their approach follows the conventional greedy approach and does not consider missing values. 

XGBoost algorithm~\citep{chen2016xgboost} also implements MIA to handle missing values. However, 
building a fair XGBoost model is not straightforward as it requires a twice-differentiable loss function while most group fairness metrics are non-differentiable. There is only one workshop paper~\citep{ravichandran2020fairxgboost} tackling this challenge, which is limited to the statistical parity metric and using a surrogate loss function. In contrast, our framework is applicable to a broad class of group discrimination metrics,  minimizes them directly in the mixed-integer program, and provides optimality guarantees.


\section{Framework}

\paragraph{Supervised learning and disparate impact.} Consider a supervised learning task where the goal is to predict an outcome random variable $Y \in \mathcal{Y}$ using an input random feature vector $X\in \mathcal{X}$. For a given loss function $\ell: \mathcal{Y}\times \mathcal{Y} \to \Reals^{+}$, the performance of an ML model $h: \mathcal{X} \to \mathcal{Y}$ can be measured by a population risk $L(h) \defined \EE{\ell(h(X), Y)}$. Some typical choices of the loss function are 0-1 loss $\ell_{0\text{-}1}(\hat{y}, y) = \indicator{[\hat{y} = y]}$ for classification task or mean squared loss $\ell_{2}(\hat{y}, y) = (\hat{y} - y)^2$ for regression. Since the underlying distribution $P_{X,Y}$ is unknown, one usually minimizes an empirical risk instead  using a dataset of samples drawn from $X$ and $Y$ given by $\mathcal{D}=\left\{(\mbx_i, y_i)\right\}_{i=1}^n$:
\begin{align}
\label{eq::min_emp_risk}
    \min_{h \in \mathcal{H}}~\frac{1}{n} \sum_{i=1}^n \ell(h(\mbx_i), y_i),
\end{align}
where $\mathcal{H}$ is a class of predictive models  (e.g., neural networks or logistic regression). 

A model $h$ exhibits disparate impact \citep{barocas2016big} if its performance varies across population groups, defined by a group attribute $S$. For simplicity, we assume that $S$ is binary, but our framework can be extended to a more general intersectional setting. For a given loss function $\ell$, we measure the performance of $h$ on group $s$ by $L_s(h) \defined \EE{\ell(h(X),Y) \mid S=s}$ and say that $h$ is discriminatory if $L_0(h) \neq L_1(h)$. We define the \emph{discrimination risk} as: 
\begin{equation}\label{eq:disc}
    \mathsf{Disc}(h) \defined |L_0(h) - L_1(h)|.
\end{equation}
We refer the readers to \citet{dwork2018decoupled,donini2018empirical} for a discussion on how to recover some commonly used group fairness measures, such as equalized odds~\citep{hardt2016equality}, by selecting different loss functions in \eqref{eq:disc}. In this paper,  we say a model ensures fairness when \eqref{eq:disc} is small or zero for a chosen loss function.

\paragraph{Data missingness.} In practice, data may contain features with missing values. We assume that the complete data are composed of two components:  observed variables $X_{\text{obs}}$ and  a missing variables $X_{\text{ms}}$. For the purpose of our theoretical analysis, we assume that only single variable is missing---a common simplifying assumption made in the statistics literature~\citep{little2019statistical}. 
We introduce a binary variable $M$ for indicating whether $X_{\text{ms}}$ is  missing (i.e., $M=1$ if and only if $X_{\text{ms}}$ is missing). Finally, we define the incomplete feature vector by $\tilde{X} = (X_{\text{obs}}, \tilde{X}_{\text{ms}}) \in \tilde{\mathcal{X}}$ where 
\begin{align*}
    \tilde{X}_{\text{ms}}
    = 
    \begin{cases}
    X_{\text{ms}}  &\text{ if } M = 0\\
    *  &\text{ otherwise.}
    \end{cases}
\end{align*}
Note that we drop the assumption that only one variable is missing when developing the Fair MIP Forest in Section \ref{sec::fair_dec_tree}.

Missing data can be categorized into three types \citep{little2019statistical} based on the relationship between missing pattern and observed variables:
\begin{itemize}
    \item Missing completely at random (MCAR) if $M$ is independent of $X$;
    \item Missing at random (MAR) if $M$ depends only on the observed variables $X_{\text{obs}}$;
    \item Missing not at random (MNAR) if neither MCAR nor MAR holds.
\end{itemize}

Even though most missing patterns in the real world are MNAR, the theoretical studies on imputation methods often rely on the MCAR (or MAR) assumption. However, when the missing pattern varies across groups, even if the each group satisfies MCAR/MAR, it is possible that entire population does not, and vice versa. This is formalized in the following lemma. 
\begin{lem}
\label{lem::missing_pattern}
There exist data distributions such that (i) each group satisfies MCAR (or MAR) but the entire population does not or (ii) the entire population satisfies MCAR (or MAR) but each group does not.
\end{lem}

\paragraph{Data imputation.} Using incomplete feature vectors for solving the empirical risk minimization in~\eqref{eq::min_emp_risk} may be challenging since most optimizers, such as gradient-based methods, only take real-valued inputs. To circumvent this issue, one can impute missing values by applying a mapping $f_{\text{imp}}: \tilde{\mathcal{X}} \to \mathcal{X}$ on the incomplete feature vector $\tilde{X}$. 


\section{Risks of Training with Imputed Data}
\label{sec::risk_train_imp}
A model trained on imputed data may manifest disparate impacts for various reasons. For example, the imputation method can be inaccurate for the minority group, leading to a
biased imputed dataset. As a result, models trained on the imputed dataset may inherit this bias. In this section, we identify three different ways imputation can fail to produce fair models: (i) the imputation method may have different performance per group, (ii) there can be a mismatch between the imputation methods used during training and testing time, and (iii) the imputation method may be applied without the knowledge of the downstream predictive task. For the last point, we prove that there is no universal imputation method that ensures fairness across all downstream learning tasks.

\subsection{Biased Imputation Method}
To quantify the performance of an imputation method and how it varies across different groups, we introduce the following definition. 
\begin{defn}
The performance of an imputation method $f_{\text{imp}}$ on group $s$ is measured by
\begin{align}
    L_s(f_{\text{imp}}) \defined \EE{\|f_{\text{imp}}(\tilde{X}) - X\|_2^2 \mid M=1,S=s}.
\end{align}
Furthermore, we define the \emph{discrimination risk} of $f_{\text{imp}}$ by 
\begin{align}
    \mathsf{Disc}(f_{\text{imp}}) \defined |L_0(f_{\text{imp}}) - L_1(f_{\text{imp}})|.
\end{align}
\end{defn}

Next, we show that even when data satisfy the simplest assumption of MCAR within each group, the optimal imputation  $f_{\text{imp}}^*$ can still exhibit discrimination risk. 
\begin{thm}
\label{thm::dec_disc_imp}
Assume that data from each group are MCAR. For the sake of illustration, we let $X_{\text{obs}} = \emptyset$ and the optimal imputation method be 
\begin{align*}
    f_{\text{imp}}^* 
    = \argmin_{f_{\text{imp}}}~\EE{(f_{\text{imp}}(\tilde{X})-X)^2\mid M=1}.
\end{align*}
We can decompose its discrimination risk as
\begin{align*}
    \mathsf{Disc}(f^*_{\text{imp}})
    = |&(p_1^{\text{ms}} - p_0^{\text{ms}}) (m_1-m_0)^2 \\
    & + (\var{X | S=0} - \var{X | S=1}) |
\end{align*}
where $p_s^{\text{ms}} \defined \Pr(S=s|M=1)$ and $m_s \defined \EE{X| S=s}$ for $s\in \{0,1\}$.
\end{thm}
This theorem reveals three factors which may cause discrimination in data imputation: (i) different proportion of missing data, $p_1^{\text{ms}} - p_0^{\text{ms}}$, (ii) difference in group means, $m_1 - m_0$, and (iii) different variance per group, $\var{X | S=0} - \var{X | S=1}$. 
The first factor could be caused by different missing pattern across groups.
Since $X_{\text{obs}} = \emptyset$, an optimal imputation method would be imputing a constant, and the optimal constant is the population mean for the L2 loss.\footnote{See Section A.2 in the supplementary material.} Hence, the second factor measures the gap between the per-group optimal imputation methods. 
Finally, the variance quantifies the performance of the mean imputation on each group. Therefore, the last factor indicates how suitable the mean imputation is for each group.

\begin{rem}
We briefly discuss how to improve the fairness of  imputation based on the factors identified in Theorem~\ref{thm::dec_disc_imp}. First, if the bias mainly comes from the difference $p_1^{\text{ms}} - p_0^{\text{ms}}$, then one can collect more samples from the minority group which has more missing data or resample the training dataset. Second, if the mean values of each group are significantly different (i.e., $m_1 - m_0$ is large), then one can impute the missing values for each group separately, assuming it is legal and ethical to do so \citep{dwork2018decoupled,wang2021split}. Lastly, if the variance difference $\var{X | S=0} - \var{X | S=1}$ is large, then one may consider using a different imputation mechanism for each group instead of mean imputation.
\end{rem}

\subsection{Mismatched Imputation Methods}

 Imputation is an unavoidable process to train many classes of ML models (e.g. logistic regression) in the presence of missing values. However, a user of a trained model might not have information on what type of imputation was performed for training. People who disseminate a trained model might omit the details about the imputation process. In some cases, it is not possible for a model developer to disclose all the details about the imputation due to privacy concerns. Take mean imputation as an example. Releasing the sample mean can leak private information in the training set, and a user might have to estimate the sample mean from the testing data, which would not precisely match that of training data. 
In the following theorem, we provide how this imputation mismatch between training and testing can aggravate the discrimination risk.

\begin{thm}
\label{thm::disc_dif_imp}
For a predictive model $h$ and an imputation method $f_{\text{imp}}$, the performance on group $s$ is measured by 
\begin{align}
    L_s(h\circ f_{\text{imp}}) \defined \EE{\ell(h\circ f_{\text{imp}}(\tilde{X}), Y) \mid S=s}.
\end{align}
Assume that the loss function $\ell$ is bounded by a constant $K>0$ and data from group $s\in \{0,1\}$ are MCAR with probability $p_s$. Let $f_{\text{imp}}^{\text{train}}$ and $f_{\text{imp}}^{\text{test}}$ be the imputation methods used in the training and testing time, respectively. Then
\begin{equation}
\label{eq:thm2}
\begin{aligned}
    &\left|L_0(h\circ f_{\text{imp}}^{\text{test}}) - L_1(h\circ f_{\text{imp}}^{\text{test}})\right| \\
    &\leq \left|L_0(h\circ f_{\text{imp}}^{\text{train}}) - L_1(h\circ f_{\text{imp}}^{\text{train}})\right| \\
    &\quad + K \sum_{s} p_s \TV(P^{\text{train}}_s \| P^{\text{test}}_s). 
\end{aligned}
\end{equation}
where $\TV(\cdot\|\cdot)$ is the total variation distance and $P_{s}^{\text{train}}$, $P_{s}^{\text{test}}$ are the probability distributions of $(f_{\text{imp}}^{\text{train}}(\tilde{X}),Y) | M=1,S=s$ and $(f_{\text{imp}}^{\text{test}}(\tilde{X}),Y) | M=1,S=s$, respectively. Finally, there exist a data distribution, a predictive model, and an imputation method such that the equality in \eqref{eq:thm2} is achieved. 
\end{thm}

The second term in the upper bound in~\eqref{eq:thm2} shows that even if discrimination risk is completely eliminated at training time, a mismatched imputation method can still give rise to an overall discrimination risk at  testing time.

\subsection{Imputation Without Being Aware of the Downstream Tasks}

Imputing missing values and training predictive model are closely intertwined. On the one hand, the performance of the predictive model relies on how missing data are imputed. On the other hand, if data are imputed blindly without taking the downstream tasks into account, the predictive model produced from the imputed data can be unfair. 
We ask a fundamental question on whether there exists a universally good imputation method that can guarantee fairness and  accuracy regardless of which model class is used in the downstream.
We address this question by borrowing notions from learning theory.

Many existing fairness intervention algorithms \citep[see e.g.,][]{donini2018empirical,zafar2019fairness,celis2019classification,wei2020optimized,alghamdi2020model} can be understood as solving the following optimization problem (or its various approximations) for a given $\epsilon \geq 0$
\begin{align}
    \min_{h\in\mathcal{H}}&~L(h) \label{eq::sub_fair_opt_gen}\\
    \sto&~|L_0(h) - L_1(h)| \leq \epsilon. \nonumber
\end{align}

Since the imputation method changes the data distribution, it affects the solution of the optimization problem in \eqref{eq::sub_fair_opt_gen} as well. The following definition characterizes the class of imputation methods that ensure the existence of an accurate solution of \eqref{eq::sub_fair_opt_gen}.
\begin{defn}
\label{defn::ideal_imp}
For a given hypothesis class $\mathcal{H}$, a distribution of $(S,\tilde{X},Y)$, and constants $\epsilon,\delta\geq 0$, we call an imputation method $(\epsilon,\delta)$-\emph{conformal} if the minimal value of \eqref{eq::sub_fair_opt_gen} under the imputed data distribution is upper bounded by $\delta$.
\end{defn}
Although the above definition does not say anything about how an optimal solution can be reached, a conformal imputation guarantees the existence of a fair solution. On the other hand, if non-conformal method is applied to impute data, solving \eqref{eq::sub_fair_opt_gen} will always return an unfair (or inaccurate) solution, no matter which optimizer is used.

By definition, whether an imputation method is conformal relies on the hypothesis class. Hence, one may wonder if there is a universally conformal imputation method that can be applied to missing values regardless of the hypothesis class. However, the following theorem states that such a method, unfortunately, does not exist.
\begin{thm}
\label{thm::no_uni_ideal_imp}
There is no universally conformal imputation method. Specifically, for $\epsilon,\delta < 0.5$, there exists a data distribution of $(S,\tilde{X},Y)$ and two hypothesis classes $\mathcal{H}_1$ and $\mathcal{H}_2$ such that their $(\epsilon,\delta)$-conformal imputation methods are disjoint.
\end{thm}
The previous theorem suggests that imputing missing values and training a predictive model cannot be treated separately. Otherwise, it may result in a dilemma where there is no fair model computed from the imputed data. On the other hand, it is often hard to find a conformal imputation method that is tailored to a particular hypothesis class. In fact, even verifying that a given imputation method is conformal is non-trivial without solving the optimization problem in \eqref{eq::sub_fair_opt_gen}. %

\section{Fair Decision Tree with Missing Values}
\label{sec::fair_dec_tree}
\begin{figure*}[t]
    \centering
    \includegraphics[width=0.8\textwidth]{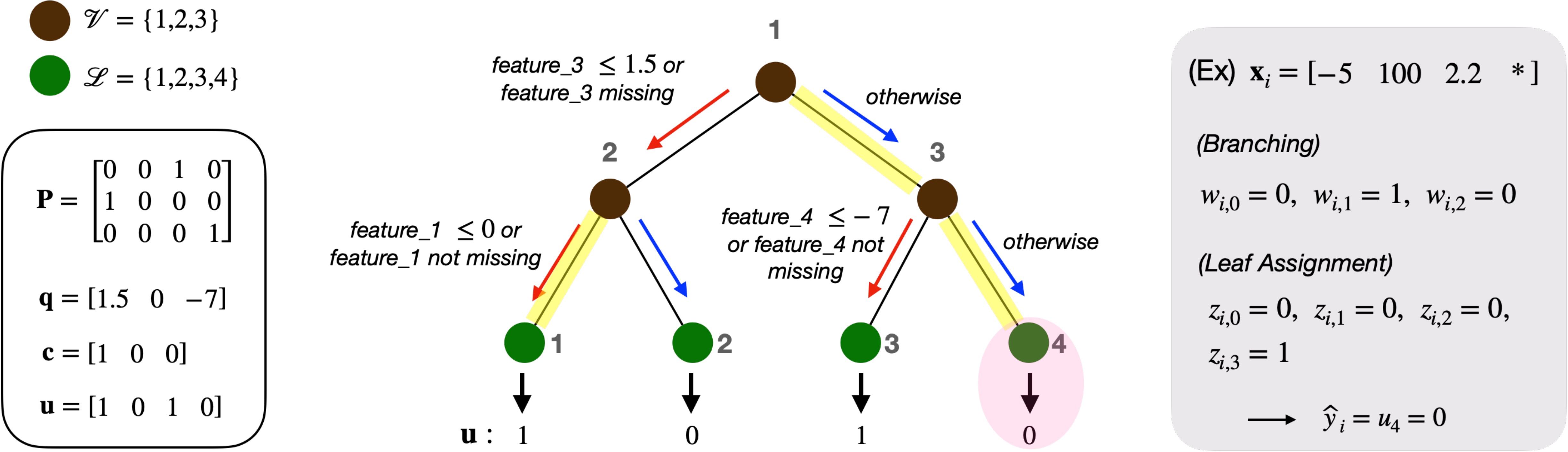}
    \caption{Demonstration of MIP notations for decision trees with MIA using a depth-2 tree example. The data dimension $d=4$, and $\mbP$ is a 3-by-4 matrix where each row dictates which feature a branching node uses for split. In the first row of $\mbP$, the third element is one, so feature\_3 is used for splitting at branch node 1. The first element in $\mbq$ and $\mbc$ are used as a splitting threshold missing values splitting at branch node 1. We also show how prediction is made for the example data point $\mbx_i = \begin{bmatrix} -5 & 100 & 2.2 & *\end{bmatrix}$. The branching decisions $w_{i,v}$'s are computed for all $v$ (highlighted in yellow). From the computed $w_{i,v}$'s, the algorithm decides the leaf node it belongs to, i.e., $\mbz_i$. In this case $\mbz_i = \begin{bmatrix} 0 & 0 & 0 & 1 \end{bmatrix}$. Hence, $\widehat{y}_i = u_4 = 0.$}
    \label{fig:MIPForest_fig}
\end{figure*}

The previous section reveals several ways discrimination risks can be introduced in the process of training with imputed data. Motivated by the limitations of decoupling imputation from model training, we propose an integrated approach for learning a fair model with missing values. The proposed approach is based on decision trees. We exploit the \emph{``missing incorporated in attribute'' (MIA)}~\cite{twala2008good}, which uses data missingness to compute the splitting criteria of a tree without performing explicit imputation. We combine MIA with the integer programming approach for training a fair decision tree. 
The details of the algorithm is described next. 

\subsection{Fair MIP Forest Algorithm} 
\label{subsec:algorithm}

In this section, we propose \emph{Fair MIP Forest} algorithm for binary classification where missing value handling is embedded in a training process that minimizes both classification error and discrimination risk. The algorithm is designed by marrying two different ideas: MIA and mixed integer programming (MIP) formulations for fitting fair decision trees. In order to mitigate the high computational cost and the overfitting risk of the integer programming method, we propose using an ensemble of under-optimized trees. We begin with providing a brief  background on MIA and the MIP formulation for  decision trees.

\paragraph{Missing Incorporated in Attribute (MIA).} 
MIA is a method that naturally handles missing values in decision trees by using missingness itself as a splitting criterion. It treats missing values as a separate category, $\{ * \}$, and at each branching node, it sends all missing values to the left or to the right. I.e., there are two ways we can split at a branch that uses the $j$-th feature: 
\begin{itemize}
    \item  $\{ X_j \leq q \text{ or } X_j = * \}$ vs $\{ X_j > q \}$, 
    \item $\{ X_j \leq q \}$ vs $\{ X_j > q \text{ or } X_j = * \}$. 
\end{itemize}
Note that by setting $q = -\infty$ in the first case, we can make the split: $\{ X_j = * \} \text{ vs }\{ X_j \neq * \}$.

\paragraph{Mixed Integer Programming (MIP) for Decision Trees.}
To set up the MIP formulation, let us introduce a few notations (see Figure~\ref{fig:MIPForest_fig} for an illustration on a depth-2 decision tree). We consider a decision tree $\calT$ of fixed depth $D$, and for simplicity we assume a full tree.  This can be described with an ordered set of branching nodes $\calV$ and an ordered set of leaf nodes $\calL$. Note that $|\calV| = 2^D - 1$ and $|\calL| = 2^D$. Learning a decision tree with missing values corresponds to learning four variables $\calT \triangleq (\mbP, \mbq, \mbc, \mbu)$: the feature we split on ($\mbP$), splitting threshold ($\mbq$), and whether to send missing values to the left or to the right  ($\mbc$) at each branching node in $\calV$, and the prediction we make at each leaf node ($\mbu$). More details on each of these variables are given below. 

Recall that $n$ is the number of samples in the training set $\calD$ and $d$ is the number of features. At each branching node $v \in \calV$, we split on one feature specified by the one-hot-encoded vector $\mbp_v \in \binset^{ d }$, and we let $\mbP \in \binset^{|\calV| \times d}$ be the matrix where each row is $\mbp_v$. Since $\mbp_v$ is a one-hot encoded vector, $\sum_{j} p_{v,j} = 1$ for all $v \in \calV$. When the value is not missing, we split at the threshold $q_v$, i.e., if the feature selected at the node $v$ is $j$, a data point that has $x_{i,j} \leq q_v$ will go to the left and a data point with $x_{i,j} > q_v$ will go to the right. When $x_{i,j}$ is missing, then it will go to the left if $c_v =1$ and to the right branch if $c_v =0$. Finally, we assume that we give the same prediction to all data points in the same leaf node, i.e., for a leaf node $l \in \calL$, the prediction will be $u_l \in \mathcal{Y}$. 

How a tree $\calT$ makes a prediction on an input data point, $(\mbx_i, y_i) \in \calD$, can be described with two more variables: $\mbw_i \in \binset^{|\calV|}$ and $\mbz_i \in \binset^{|\calL|}$ ($i \in [n]$). $\mbw_i$ represents where the data point goes to at each node, i.e., $w_{i,v} = 1$ means that the data point goes to the left branch at the branching node $v$. 
$\mbz_i$ is an one-hot-encoding vector that represents the final destination leaf node of the data point, i.e., $z_{i,l} = 1$ means that the data point goes to the leaf node $l \in \calL$, and we assign $\widehat{y}_i = u_l$. %

Under this setting, we minimize the following fairness-regularized objective function:  
\begin{equation}\label{eq:objective}
    \ell (\mathcal{D}) + \lambda \cdot \ell_\text{fair} (\mathcal{D}).
\end{equation}
We use 0-1 loss for $\ell (\mathcal{D})$ and implement four different $\ell_\text{fair}$: accuracy difference, FNR difference, FPR difference, and  equalized odds. Section B in the supplementary material describe in detail how we encode these into a mixed integer linear program. 

\paragraph{Fair MIP Forest.} We now describe our Fair MIP Forest algorithm. We take an ensemble approach to train multiple trees from different mini-batches, and employ early termination. Solving an integer program is a computationally intensive and time-consuming process. However, we observed that in the early few iterations, we can find a reasonably well-performing model, but it takes a long time to get to the (local) optimum, making minute improvements over iterations. Based on this observation, we terminate the optimization for training each tree early after some time limit ($t_\text{limit}$), and we use multiple weakly-optimized trees. We observed that this not only reduces training time but also improves prediction accuracy. Termination time ($t_\text{limit}$) and the number of trees ($n_\text{tree}$) are hyperparameters that can be tuned. Furthermore, we initialize each tree with the tree obtained in the previous iteration, to reduce the time required in the initial search. 
After we obtain $\{\calT^{(i)} \}_{i = 1, \ldots, n_\text{tree}} $, for prediction on a new data point, we perform prediction on each tree and  make a final decision using the majority rule.



\section{Experimental Results}

We present the implementation of the Fair MIP Forest algorithm. Through experiments on three datasets, we compare our proposed algorithm and existing fair learning algorithms coupled with different imputation methods.

\begin{figure}
     \centering
     \begin{subfigure}[b]{0.45\textwidth}
         \centering 
         \includegraphics[width=\textwidth]{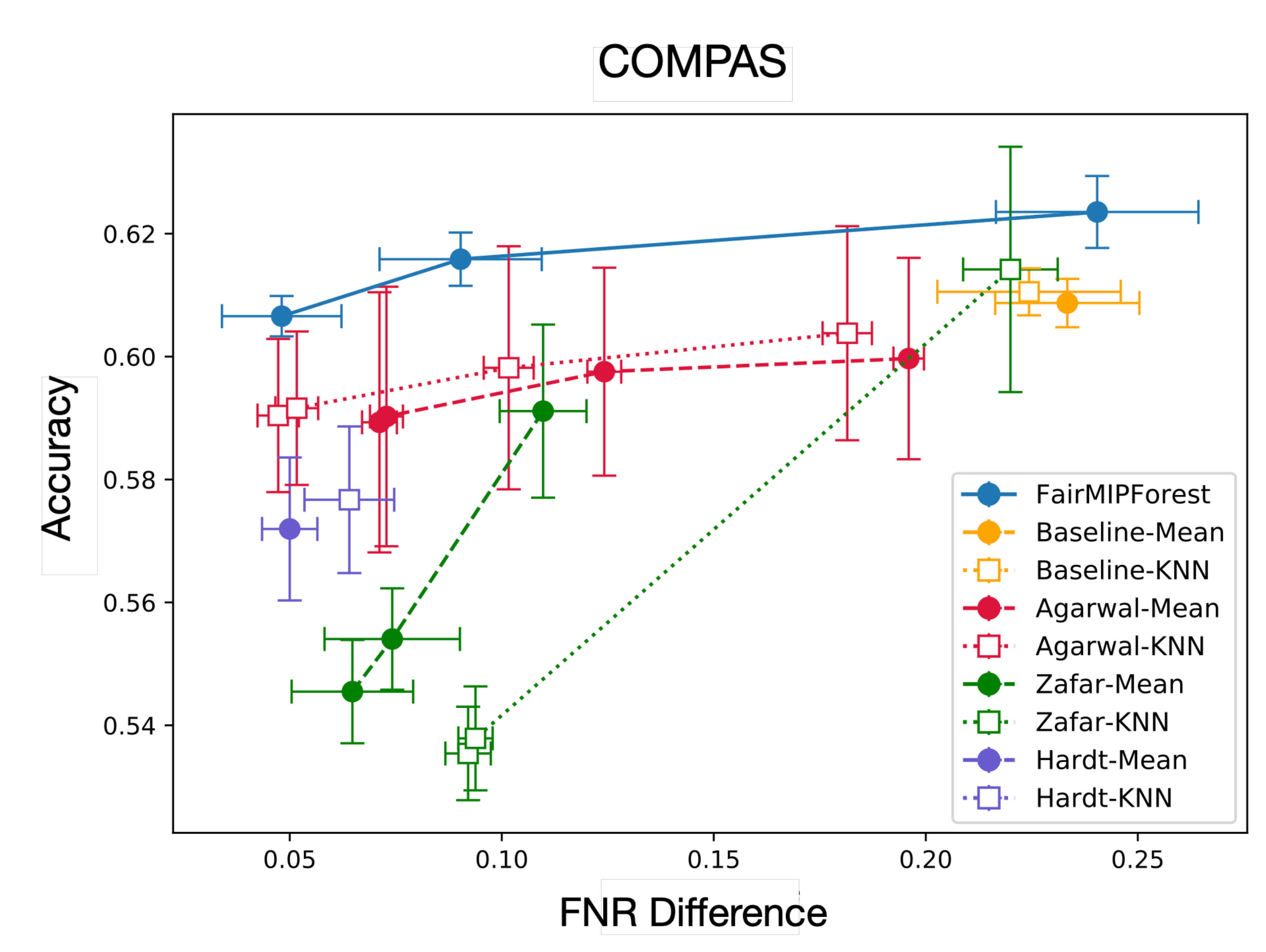}
         \caption{COMPAS}
         \label{fig:y equals x}
     \end{subfigure}
     \hfill
     \begin{subfigure}[b]{0.45\textwidth}
         \centering
         \includegraphics[width=\textwidth]{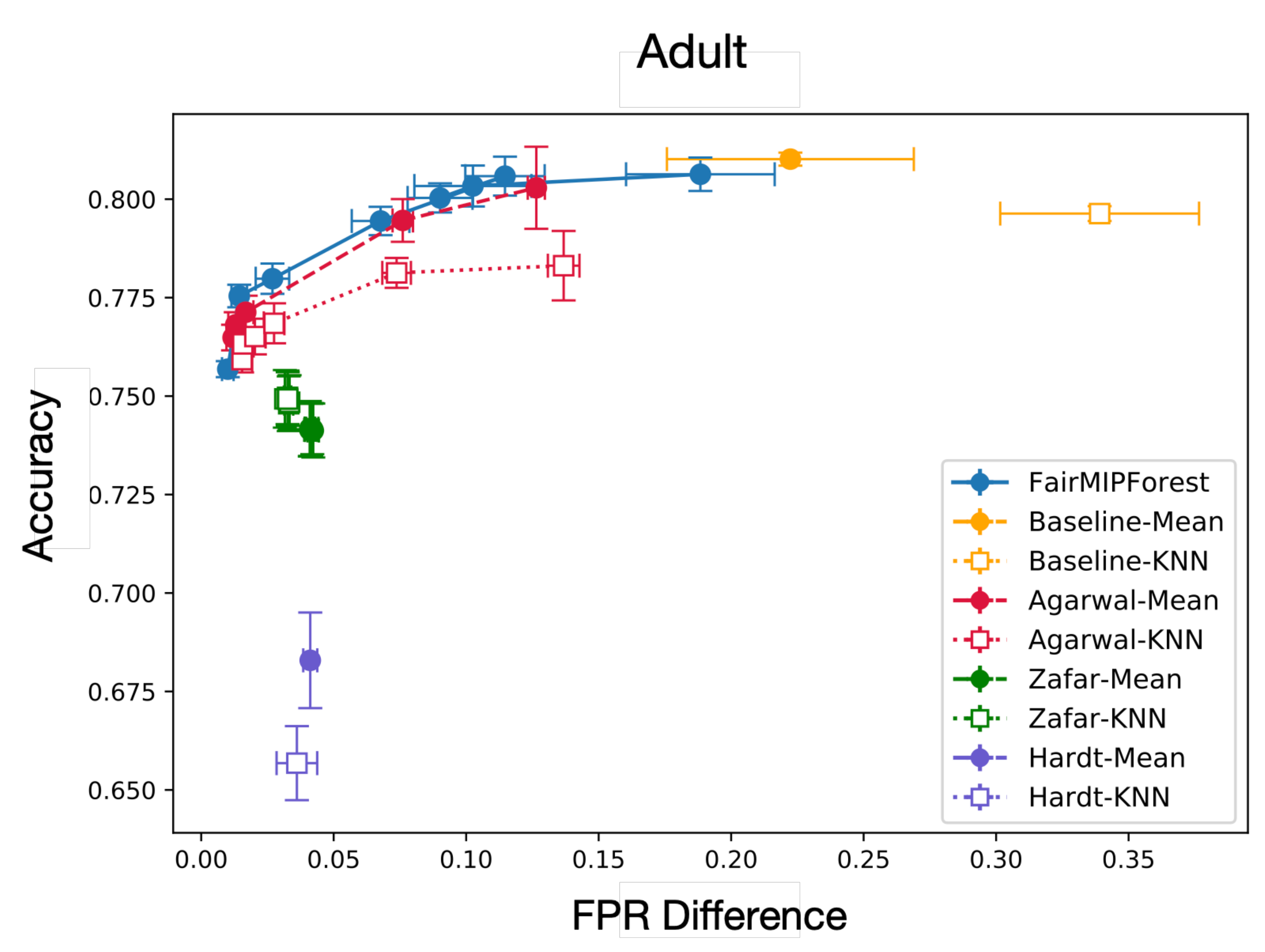}
         \caption{Adult}
         \label{fig:three sin x}
     \end{subfigure}
     \hfill
     \begin{subfigure}[b]{0.45\textwidth}
         \centering
         \includegraphics[width=\textwidth]{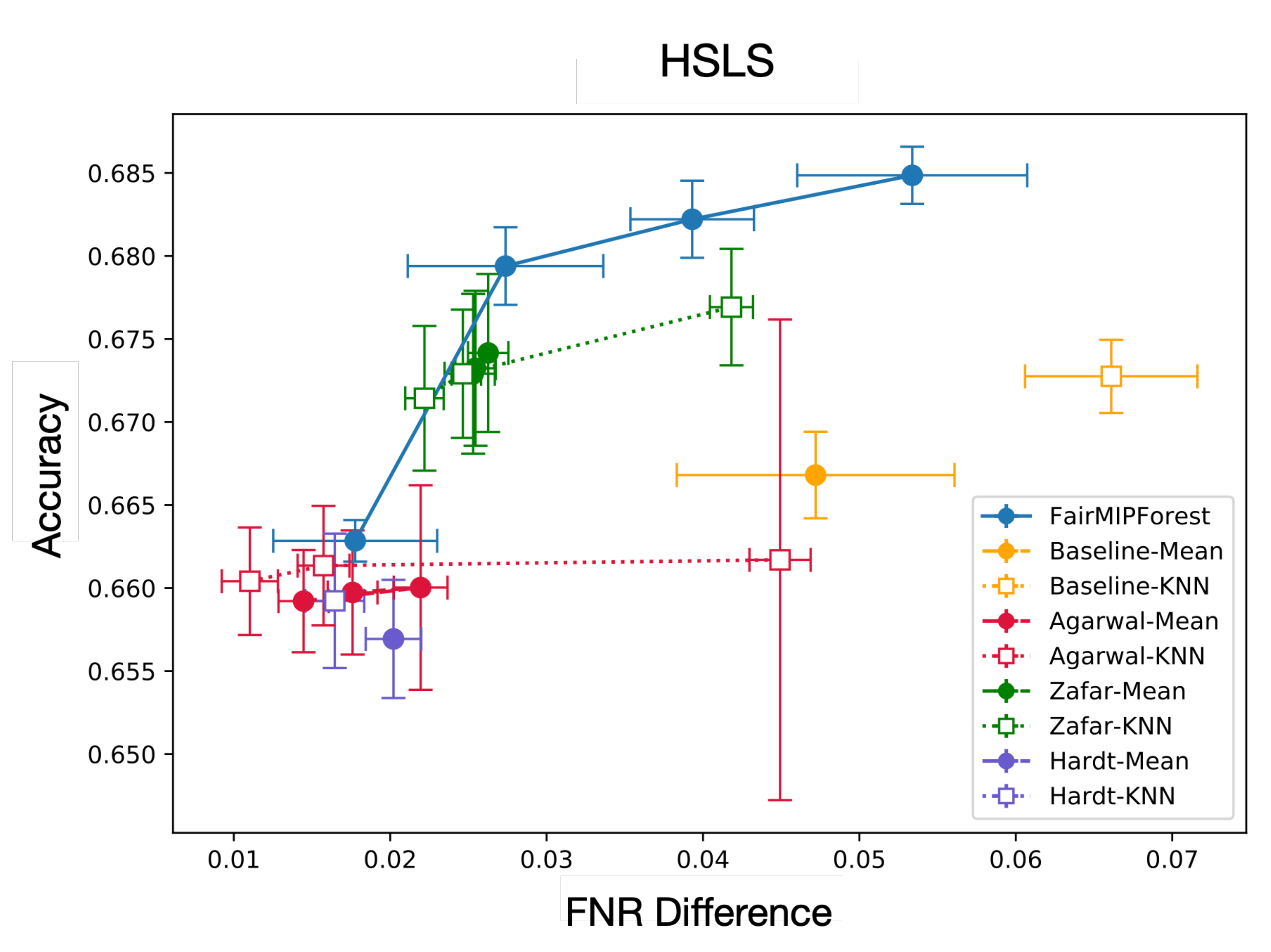}
         \caption{HSLS}
         \label{fig:five over x}
     \end{subfigure}
        \caption{Comparison of Fair MIP Forest with existing fairness intervention approaches~\cite{zafar2019fairness,hardt2016equality,agarwal2018reductions} coupled with mean or k-NN imputations. Baseline indicates the result of training a decision tree without any fair intervention. Error bars show the standard error after 10 runs with different train-test splits. }
        \label{fig:experiment}
\end{figure}

\paragraph{Datasets.} We test our Fair MIP Forest algorithm on three datasets: COMPAS~\citep{Angwin2016}, Adult~\citep{Dua:2019}, and high school longitudinal study (HSLS) dataset~\citep{ingels2011high}. For Adult and COMPAS datasets, we generate artificial missing values as the original datasets do not contain any missing values. For missing value generation, we chose a set of features to erase and for a given feature, we erased randomly with different missing probabilities per group,  which are chosen so that the baseline results (i.e., no fairness intervention) have enough disparities either in FPR or FNR (see Section C in the supplementary material).  For the HSLS dataset, there was no artificial missing value generation as the original dataset already contains a substantial amount of missing values. The sensitive attribute in Adult dataset was gender (0: Female, 1: Male), in COMPAS dataset was race (0: Black, 1: White), and in HSLS was also race (0:URM, 1:White/Asian).

\paragraph{Setup.} 
We implemented the the Fair MIP Forest algorithm with Gurobi~9.1.2 for solving the integer program. All experiments were run on a cluster with 32 CPU cores and 32GB memory. 
For each dataset, we tune hyperparameters for the Fair MIP Forest algorithm: tree depth, number of trees, batch size, time limit for each tree training (see Section C in the supplementary material for details). We vary $\lambda$ in \eqref{eq:objective} to obtain different points on the fairness-accuracy curve. We compare Fair MIP Forest with the baseline, which performs simple imputation (mean or k-NN) and then trains a decision tree. We compare our method with three fair learning methods coupled with imputation: the exponentiated gradient algorithm~\citep{agarwal2018reductions}, disparate mistreatment algorithm~\citep{zafar2019fairness}, and equalized odds algorithm~\citep{hardt2016equality}. We refer to these by the first author's name. 
All experiments are run 10 times with different random train-test splits to measure the variance in both accuracy and fairness metrics. For COMPAS and HSLS datasets, we choose FNR difference as a fairness metric because the difference in FPR was already very small in the baseline classifier without any fairness intervention. Similarly, we only regularize FPR difference in Adult dataset as FNR difference in the baseline was negligible. For Agarwal and Zafar, we vary hyperparameters to get different fairness-accuracy trade-off. Hardt does not have tunable hyperparameters.

\paragraph{Discussion.} The experimental results are summarized in Figure~\ref{fig:experiment}.  
With the COMPAS dataset, we observe that the Fair MIP Forest algorithm achieves a better accuracy for the same FNR difference, compared to all the other existing works we tested. In the Adult dataset, Fair MIP Forest and Agarwal coupled with mean imputation showed the best performance. In the HSLS experiments, Fair MIP Forest showed superior accuracy for higher FNR difference values although Agarwal was able to achieve smaller FNR difference (close to 0.01).

We also observe that even with the same dataset, there is no one imputation method that has better performance in all fair learning algorithms. For example, in Adult, while Agarwal clearly performs better with mean imputation, Zafar has slightly better fairness and accuracy with k-NN imputation. Also, notice that the performance of a fair learning algorithm depends heavily on which imputation method was used. In the  COMPAS dataset, Zafar with mean imputation performs significantly better than Zafar with k-NN imputation.
In the Adult dataset, Agarwal with mean imputation had better performance than Agarwal with k-NN. 
This suggests that how to pair which imputation with which fair learning method for a given dataset is not a straightforward question. With the Fair MIP Forest algorithm, we can sidestep  such question and simply apply it to any dataset with missing values.

For all experiments, we set $t_\text{limit}$ to 60 seconds, which means that training 30 trees takes roughly 1,800 seconds. Compared to what is reported in \citet{aghaei2019learning} -- 15,000+ seconds for training a tree for COMPAS and Adult datasets -- this is more than 8x time saving.

\section{Discussion and Future Work} \label{sec:discussion}
In this work, we analyze different sources of fairness risks, in terms of commonly-used group fairness metrics, when we train a model from imputed data.
Extension of our analysis to multiple imputation (e.g., MICE~\citep{van2011mice}) and to other fairness metrics (e.g., individual fairness~\citep{dwork2012fairness}, preference-based fairness~\citep{zafar2017}, or rationality~\citep{ustun2019fairness}) would be an interesting future direction. 
We then introduce our solution to training a fair model with missing values, that utilizes decision trees. While tree-based algorithms are a preferred choice in many settings for their interpretability and ability to accommodate mixed data types (categorical and real-valued), we hope our work can inspire the development of fair handling of missing values in other supervised models, such as neural networks. 
Finally, the general question of how to design a fair imputation procedure is a widely open research problem. 

\section{Acknowledgement}
This material is based upon work supported by the National Science Foundation under grants CAREER 1845852, IIS 1926925, and FAI 2040880.

\clearpage

\bibliography{references}

\clearpage
\onecolumn
\appendix
\section{Omitted Proofs}
\subsection{Proof of Lemma~\ref{lem::missing_pattern}}
\begin{proof}
We assume that $X_{\text{obs}} = \emptyset$. In this case, MCAR and MAR are equivalent so we only focus on MCAR in what follows. First, we construct a probability distribution such that each group satisfies MCAR but the entire population does not. Let $S\sim \text{Bernoulli}(0.5)$ and
\begin{align*}
    &X|S=0 \sim \text{Bernoulli}(0.1),\quad \ X|S=1 \sim \text{Bernoulli}(0.9),\\
    &M|S=0 \sim \text{Bernoulli}(0.1), \quad M|S=1 \sim \text{Bernoulli}(0.9).
\end{align*}
By construction, we know each group satisfies MCAR. However,
\begin{align*}
    &\Pr(M=1, X=1)\\
    &= \sum_{s} \Pr(M=1, X=1|S=s) \Pr(S=s)\\
    &= \sum_{s} \Pr(M=1|S=s)\Pr(X=1|S=s) \Pr(S=s)\\
    &= 0.1 \times 0.1 \times 0.5 + 0.9 \times 0.9 \times 0.5
    = 0.41,
\end{align*}
and
\begin{align*}
    &\Pr(M=1) \Pr(X=1)\\
    &= \sum_{s} \Pr(M=1|S=s) \Pr(S=s) \sum_s \Pr(X=1|S=s) \Pr(S=s)\\
    &= (0.1 \times 0.5 + 0.9 \times 0.5) \times (0.1 \times 0.5 + 0.9 \times 0.5)
    = 0.25.
\end{align*}
Hence, $\Pr(M=1, X=1) \neq \Pr(M=1) \Pr(X=1)$ which means that the entire population does not satisfy MCAR. 

Next, we construct a probability distribution such that the entire population satisfies MCAR but each group does not. Let $S\sim \text{Bernoulli}(0.5)$, 
\begin{align*}
    &\Pr(M=0, X=0|S=0) = 0.1, \quad \Pr(M=0,X=1|S=0) = 0.3,\\
    &\Pr(M=1, X=0|S=0) = 0.4, \quad \Pr(M=1,X=1|S=0) = 0.2,
\end{align*}
and
\begin{align*}
    &\Pr(M=0, X=0|S=1) = 0.4, \quad \Pr(M=0,X=1|S=1) = 0.2,\\
    &\Pr(M=1, X=0|S=1) = 0.1, \quad \Pr(M=1,X=1|S=1) = 0.3.
\end{align*}
By construction, 
\begin{align*}
    &\Pr(M=0, X=0) = 0.25, \quad \Pr(M=0,X=1) = 0.25,\\
    &\Pr(M=1, X=0) = 0.25, \quad \Pr(M=1,X=1) = 0.25.
\end{align*}
As a result, the entire population satisfies MCAR but each group does not satisfy this assumption.
\end{proof}

\subsection{Proof of Theorem~\ref{thm::dec_disc_imp}}
\begin{proof} 
We denote $\alpha \defined f_{\text{imp}}(*)$ and let $p_s^{\text{ms}} \defined \Pr(S=s|M=1)$, $m_s \defined \EE{X| S=s}$ for $s\in \{0,1\}$.
Then
\begin{align*}
    &\EE{(f_{\text{imp}}(\tilde{X}) - X)^2 \mid M=1}\\
    &= \sum_{s} \Pr(S=s| M=1) \EE{(\alpha - X)^2 \mid M=1,S=s}\\
    &= \sum_{s} p_s^{\text{ms}}\EE{(\alpha - X)^2 \mid S=s}.
\end{align*}
Consequently, we have
\begin{align*}
    \alpha^*
    &=\argmin_{\alpha}~\sum_{s} p_s^{\text{ms}} \EE{(\alpha - X)^2 \mid S=s}\\
    &= \argmin_{\alpha}~\sum_{s} p_s^{\text{ms}}  (\alpha-m_s)^2\\
    &= \frac{p_0^{\text{ms}} m_0 + p_1^{\text{ms}} m_1}{p_0^{\text{ms}} + p_1^{\text{ms}}}\\
    &=p_0^{\text{ms}} m_0 + p_1^{\text{ms}} m_1.
\end{align*}
Therefore, the optimal imputation method is unique and has a closed-form expression:
\begin{align*}
    f_{\text{imp}}(\tilde{x})
    = \begin{cases}
    \tilde{x} \quad &\text{if }\ \tilde{x}\in \mathcal{X}\\
    \alpha^* \quad &\text{if }\ \tilde{x} = *. 
    \end{cases}
\end{align*}
The performance of the optimal imputation method on group $s$ is
\begin{align*}
    \EE{(f^*_{\text{imp}}(\tilde{X}) - X)^2\mid M=1, S=s}
    &=\EE{(\alpha^* - X)^2\mid S=s}\\
    &= \left(p_0^{\text{ms}} m_0 + p_1^{\text{ms}} m_1 - m_s\right)^2 + \var{X | S=s}\\
    &=(m_1-m_0)^2(p_{1-s}^{\text{ms}})^2 + \var{X | S=s}.
\end{align*}
Finally, we can compute the discrimination risk of the optimal imputation method:
\begin{align*}
    &\left|\EE{(f^*_{\text{imp}}(\tilde{X}) - X)^2\mid M=1, S=0}- \EE{(f^*_{\text{imp}}(\tilde{X}) - X)^2\mid M=1, S=1}\right|\\
    &=\left|(p_1^{\text{ms}} - p_0^{\text{ms}}) (m_1-m_0)^2 + (\var{X | S=0} - \var{X | S=1}) \right|.
\end{align*}
\end{proof}

\subsection{Proof of Theorem~\ref{thm::disc_dif_imp}}
\begin{proof}
Since data from each group are MCAR, the quantity $L_s(h\circ f_{\text{imp}}^{\text{test}})$ is equal to
\begin{align}
    &\EE{\ell(h\circ f_{\text{imp}}^{\text{test}}(\tilde{X}), Y) \mid M= 1, S=s} p_s + \EE{\ell(h\circ f_{\text{imp}}^{\text{test}}(\tilde{X}), Y) \mid M= 0, S=s} (1-p_s) \nonumber\\
    &= \EE{\ell(h\circ f_{\text{imp}}^{\text{test}}(\tilde{X}), Y) \mid M= 1, S=s} p_s + \EE{\ell(h(X), Y) \mid S=s} (1-p_s). \label{eq::Ls_equiv_perf}
\end{align}
Now we can rewrite the first term as 
\begin{align}
    \EE{\ell(h\circ f_{\text{imp}}^{\text{test}}(\tilde{X}), Y) \mid M= 1, S=s}
    = \int \ell(h(x), y) \dif P^{\text{test}}_s(x,y)  \label{eq::Ls_equiv_perf_sec}
\end{align}
where $P^{\text{test}}_s$ is the probability distribution of $(f_{\text{imp}}^{\text{test}}(\tilde{X}),Y) | M=1,S=s$. Combining \eqref{eq::Ls_equiv_perf} with \eqref{eq::Ls_equiv_perf_sec} yields
\begin{align}
\label{eq::Ls_eq_exp}
    L_s(h\circ f_{\text{imp}}^{\text{test}}) 
    =p_s \int \ell(h(x), y) \dif P^{\text{test}}_s(x,y) + \EE{\ell(h(X), Y) \mid S=s} (1-p_s).
\end{align}
Similarly, we have
\begin{align}
    L_s(h\circ f_{\text{imp}}^{\text{train}})
    = p_s \int \ell(h(x), y) \dif P^{\text{train}}_s(x,y)  + \EE{\ell(h(X), Y) \mid S=s} (1-p_s)
\end{align}
where $P^{\text{train}}_s$ is the probability distribution of $(f_{\text{imp}}^{\text{train}}(\tilde{X}),Y) | M=1,S=s$. Since the loss function is bounded between $0$ and $K$, the variational representation of total variation distance \citep[see Section~6.3 in][]{polyanskiy2014lecture} implies
\begin{align}
\label{eq::TV_vari}
    \left|\int \ell(h(x), y) \dif P^{\text{train}}_s(x,y) - \int \ell(h(x), y) \dif P^{\text{test}}_s(x,y)\right|
    \leq K \TV(P^{\text{train}}_s \| P^{\text{test}}_s).
\end{align}
By the triangle inequality and (\ref{eq::Ls_eq_exp}--\ref{eq::TV_vari}), we have
\begin{align}
\label{eq::ineq_shift}
    \left|L_0(h\circ f_{\text{imp}}^{\text{test}}) - L_1(h\circ f_{\text{imp}}^{\text{test}})\right|
    \leq \left|L_0(h\circ f_{\text{imp}}^{\text{train}}) - L_1(h\circ f_{\text{imp}}^{\text{train}})\right| + K \sum_{s} p_s \TV(P^{\text{train}}_s \| P^{\text{test}}_s).
\end{align}
Finally, we prove that the above inequality is tight. Let the loss function be the 0-1 loss and $X = (X_{\text{obs}}, X_{\text{ms}}) \in [0,1]^2$, $Y \in \{0,1\}$. Consider a binary classifier $h(x_1,x_2) = \indicator{[x_2 \geq 0.5]}$ and different imputation methods during training and testing time: $f_{\text{imp}}^{\text{train}}(x_{\text{obs}},*) = (x_{\text{obs}}, 0)$ and $f_{\text{imp}}^{\text{test}}(x_{\text{obs}},*) = (x_{\text{obs}}, \indicator{[x_{\text{obs}} \leq 0.5]})$. Furthermore, we let the missing probability $p_s = 1$ and $P_{\tilde{X},Y|M=1,S=0} = \delta_{(0,*),0}$, $P_{\tilde{X},Y|M=1,S=1} = \delta_{(1,*),0}$. In this case, $L_0(h\circ f_{\text{imp}}^{\text{train}}) = L_1(h\circ f_{\text{imp}}^{\text{train}}) = 0$, $L_0(h\circ f_{\text{imp}}^{\text{test}}) = 1$, $L_1(h\circ f_{\text{imp}}^{\text{test}}) = 0$. Consequently, the LHS and RHS of \eqref{eq::ineq_shift} are both $1$.
\end{proof}

\subsection{Proof of Theorem~\ref{thm::no_uni_ideal_imp}}
\begin{proof}
Let the loss function used for defining $L(h)$ and $L_s(h)$ be the 0-1 loss and $X = (X_{\text{obs}}, X_{\text{ms}}) \in [0,1]^2$, $Y \in \{0,1\}$. Furthermore, we let $\Pr(S=0) = 0.5$, $\Pr(M=1) = 1$, and $P_{\tilde{X},Y|M=1,S=0} = \delta_{(0,*),1}$, $P_{\tilde{X},Y|M=1,S=1} = \delta_{(1,*),0}$. Now consider two hypothesis classes: 
\begin{align*}
    \mathcal{H}_1 
    \defined \{\indicator{[x_1\geq a]} \mid a\in [0,1]\},\\
    \mathcal{H}_2
    \defined \{\indicator{[x_2 \geq a]} \mid a\in [0,1]\}.
\end{align*}
For $\epsilon<0.5$, the minimum in \eqref{eq::sub_fair_opt_gen} under the imputed data distribution and hypothesis class $\mathcal{H}_1$ is always $1.0$ no matter which imputation method is used. In other words, for $\epsilon,\delta<0.5$, the class of $(\epsilon,\delta)$-conformal imputation methods under $\mathcal{H}_1$ is empty. However, for another hypothesis class, $f_{\text{imp}}(x_{\text{obs}},*) = (x_{\text{obs}}, \indicator{[x_{\text{obs}} \leq 0]})$ is a $(\epsilon,\delta)$-conformal imputation method and $h(x_1,x_2) = \indicator{[x_2 \geq 0.5]} \in \mathcal{H}_2$ gives a perfect solution of \eqref{eq::sub_fair_opt_gen}.

\end{proof}
\section{Mixed Integer Programming for Fair Decision Tree with Missing Values}\label{app:optimization}

The full integer program for training a fair decision tree with MIA is given in Program~\ref{prog:fair_tree}. We first explain the variables and then walk through each constraint in the program. 

As explained in Section~\ref{subsec:algorithm}, $p_{v,j}, q_v, c_v, u_l$ are parameters that determine the tree ($v \in \calV, j \in [d], l \in \calL$). $w_{i,v}, w_{i,v}^{(1)}, w_{i,v}^{(2)}, w_{i,v}^{(\text{nm})}$ and $z_{i,l}$ are parameters associated with the training data point $(\mbx_i, y_i)$.  $w_{i,v} \in \binset$ determines whether the $i$-th data point goes to the left or to the right at the branching node $v$. This is computed for all $v \in \calV$ regardless of whether the node is on the path to its destination leaf node or not. $w_{i,v}^{(1)}, w_{i,v}^{(2)}, w_{i,v}^{(\text{nm})}$ are auxiliary variables for computing $w_{i,v}$. 
$\mbz_i$ is an one-hot-encoded vector of length $|\calL|$ that encodes the destination leaf node of the $i$-th data point, i.e., $z_{i,l}=1$ indicates that the leaf $l$ is the destination leaf node for the $i$-th data point. $f_{l,0}, f_{l,1}$, and $\text{loss}(l)$ are variables used to compute  $\ell(\calD)$  and $\ell_\text{fair}(\calD)$ in the objective. 

The first constraint in \eqref{eq:one_hot_const} enforces the one-hot-encoding of $\mbp_v$ and $\mbz_i$. Constraints in \eqref{eq:branch_const1}--\eqref{eq:branch_const6} are used to obtain $w_{i,v}$. \eqref{eq:branch_const1} and \eqref{eq:branch_const2} encode the following logical constraint:
\begin{equation*}
    w_{i,v}^{\text{(nm)}} = \begin{cases} 1, \;\; \text {if } \; q_v \geq \sum_{j \in [d]} p_{v,j} (1-m_{i,j}) x_{i,j}, \\
    0, \;\; \text{otherwise.}
    \end{cases}
\end{equation*}
$M \in \mathbb{R}$ in \eqref{eq:branch_const1} is a constant chosen to be large enough so that the left hand side (LHS) is always smaller than $M$, and $\epsilon$ is a small constant to close to zero (e.g., 0.001). Having $\epsilon$ allows for numerical errors in the real-number representation of binary variables. Notice that $w_{i,v}^{\text{(nm)}}$ is $1$ when the selected feature at node $v$ is not missing and smaller than the threshold $q_v$. However, when the feature is missing, the right hand side (RHS) is always zero. Then, $w_{i,v}^{\text{(nm)}}$ becomes equivalent to a logical variable that represents if $q_v \geq 0$. This is an condition does not say anything about where the given data point should go, i.e., the condition encoded by $w_{i,v}^{\text{(nm)}}$ is relevant only when the value is not missing. We introduce another variable $w_{i,v}^{(1)}$ that is $1$ only if $w_{i,v}^{\text{(nm)}} = 1$ and also the value is not missing: 
\begin{equation*}
    w_{i,v}^{(1)} = \begin{cases}
    1, \;\; \text{if } \left(\sum_{j \in [d]} p_{v,j} \cdot (1-m_{i,j}) > 0 \right) \; \text{ AND } \; \left( w_{i,v}^{\text{(nm)}}>0 \right),\\ 
    0, \;\; \text{otherwise,}
    \end{cases}
\end{equation*}
and this is obtained through constraints \eqref{eq:branch_const3} and \eqref{eq:branch_const4}. When the feature is missing, we use $w_{i,v}^{(2)}$ to determine the splitting:
\begin{equation*}
    w_{i,v}^{(2)} = \begin{cases}
    1, \;\; \text{if } \left( \sum_{j \in [d]} p_{v,j} \cdot m_{i,j} > 0 \right) \; \text{ AND } \; \left( c_v > 0 \right) ,\\ 
    0, \;\; \text{otherwise,}
    \end{cases}
\end{equation*}
and this is computed through line \eqref{eq:branch_const5}. Finally, $w_{i,v}$ can be obtained from: 
\begin{equation*}
    w_{i,v} =  w_{i,v}^{(1)} \; \text{ OR } \; w_{i,v}^{(2)} \quad \text{(line \eqref{eq:branch_const6} in Program~\ref{prog:fair_tree})}.
\end{equation*}
This can be interpreted as: the data point $(\mbx_i, y_i)$ will follow the left branch if the value is not missing and satisfies the threshold to go to the left ($x_i \leq q_{v}$), or if the value is missing and $c_v =1$. Otherwise, it goes to the right branch. When we have $w_{i,v}$ for all $v \in \calV$, $z_{i,l}$ can be determined through the conditions given in \eqref{eq:branch_const7}.

As we consider binary classification problem, we choose $u_l$, prediction at leaf $l$,  based on the majority rule: 
\begin{gather*}
    u_l = \begin{cases}
    1, \quad \text{ if } \sum_{i \in [n]} y_i z_{i,l} \geq \sum_{i \in [n]} (1-y_i) z_{i,l}, \\
    0, \quad \text{ otherwise.}
    \end{cases}
\end{gather*}
This is modeled into constraints \eqref{eq:leaf_const1},\eqref{eq:leaf_const2}. For the loss function $\ell(\calD)$, we use 0-1 loss, $\ell (\mathcal{D}) = \frac{\sum_{l \in \calL} \text{loss}(l)}{n}$, where
        \begin{align}
            \text{loss}(l) = \begin{cases}
            \sum_{i \in [n]} (1-y_i) z_{i,l}, \quad \text{ if } u_l = 1,  \\ 
            \sum_{i \in [n]} y_i z_{i,l}, \quad \text{ if } u_l = 0.
            \end{cases} \label{eq:ft_loss}
        \end{align}
Since this is nonlinear with respect to the variables, we model it through constraints given in \eqref{eq:loss_const1},\eqref{eq:loss_const2}. 

For the fairness regularizer, we use group fairness metrics based on confusion matrix: FNR difference, FPR difference, equalized odds (i.e., both FNR and FPR difference), and accuracy difference. We first describe how to implement accuracy difference (i.e., $| \Pr(Y \neq \widehat{Y} | S = 0) - \Pr(Y \neq \widehat{Y} | S = 1) | $ as $\ell_\text{fair}$).
To compute this, we introduce auxiliary variables $f_{l,0}$  and $f_{l,1}$  ($l \in \calL$) that denote the number of misclassified points per leaf, for group 0 and group 1, respectively. They can be written as:
    \begin{align*}
        f_{l,0} = \begin{cases}
        \sum_{i \in [n]} (1-y_i) (1-s_i) z_{i,l},  \quad \text { if } u_l = 1, \\ 
        \sum_{i \in [n]} y_i (1-s_i) z_{i,l},  \quad \text { if } u_l = 0.
        \end{cases}
    \end{align*}
    \begin{align*}
        f_{l,1} = \begin{cases}
        \sum_{i \in [n]} (1-y_i) s_i z_{i,l},  \quad \text { if } u_l = 1, \\ 
        \sum_{i \in [n]} y_i s_i z_{i,l},  \quad \text { if } u_l = 0.
        \end{cases}
    \end{align*}
$f_{l,0}$'s are computed through \eqref{eq:fair_reg_const_1}--\eqref{eq:fair_reg_const_3} and $f_{l,1}$'s are computed through \eqref{eq:fair_reg_const_4}--\eqref{eq:fair_reg_const_6}. Then, the accuracy difference between group 0 and group 1 is given as: 
    \begin{align}
        \ell_\text{fair} (\mathcal{D}) = \left| \frac{\sum_{l \in \calL} f_{l,0}}{\sum_{i \in [n]} (1-s_i) } - \frac{\sum_{l \in \calL} f_{l,1}}{\sum_{i \in [n]} s_i } \right| \label{eq:l_fair}
    \end{align}
Although the term inside the absolute value is a linear combination of the variables, taking the absolute value makes this non-linear. Hence, instead of having \eqref{eq:l_fair} in the objective function directly, we model this into constraints \eqref{eq:l_fair_1} and \eqref{eq:l_fair_2}. 

Now we describe how this can be modified to regularize to FPR or FNR difference. To use FPR difference, we set the first terms in \eqref{eq:fair_reg_const_3}, \eqref{eq:fair_reg_const_6} to zero, i.e., 
    \begin{gather*}
        f_{l,0} \geq 0 - M u_l, \;\; f_{l,1} \geq 0 - M u_l, \quad \forall l \in \calL  \\
        f_{l,0} \leq 0 + M u_l + \epsilon, \;\; f_{l,1} \leq 0 + M u_l + \epsilon, \quad \forall l \in \calL.
    \end{gather*}
Additionally, we modify $\ell_\text{fair}$ to: 
\begin{align*}
    \ell_\text{fair} (\mathcal{D}) = \left| \frac{\sum_{l \in \calL} f_{l,0}}{\sum_{i \in [n]} (1-s_i) (1-y_i) } - \frac{\sum_{l \in \calL} f_{l,1}}{\sum_{i \in [n]} s_i (1-y_i)} \right| 
\end{align*}
Similarly, to use FNR difference as a regularizer, we set the first terms in \eqref{eq:fair_reg_const_1}, \eqref{eq:fair_reg_const_2}, \eqref{eq:fair_reg_const_4},\eqref{eq:fair_reg_const_5} to zero:
\begin{gather*}
    f_{l,0} \geq 0 - M (1-u_l), \;\; f_{l,1} \geq 0 - M (1-u_l), \quad \forall l \in \calL,  \\
    f_{l,0} \leq 0 + M (1-u_l) + \epsilon, \;\;  f_{l,1} \leq 0 + M (1-u_l) + \epsilon, \quad \forall l \in \calL, and 
\end{gather*}
set $\ell_\text{fair}$ to: 
\begin{align*}
    \ell_\text{fair} (\mathcal{D}) = \left| \frac{\sum_{l \in \calL} f_{l,0}}{\sum_{i \in [n]} (1-s_i) y_i } - \frac{\sum_{l \in \calL} f_{l,1}}{\sum_{i \in [n]} s_i y_i} \right| 
\end{align*}
To use equalized odds as a regularizer, we can use above formulas to compute the FPR difference and FNR difference separately, and regularize: 
$$| \Pr(Y \neq \hat{Y} | S = 0, Y=0) - \Pr(Y \neq \hat{Y} | S = 1, Y=0) | +  | \Pr(Y \neq \hat{Y} | S = 0, Y=1) - \Pr(Y \neq \hat{Y} | S = 1,Y=1) |.$$ 
Major differences in our formulation from the previous works \citep{bertsimas2017optimal,aghaei2019learning} are:
\begin{itemize}
    \item We implement fairness regularizers that are based on the performance difference between two groups -- FNR difference, FPR difference, equalized odds, and accuracy difference -- as compared to \citet{aghaei2019learning}, which only considered statistical parity. We add leaf-wise fairness risk variables $f_{l,0}$ and $f_{l,1}$ to implement this. 

    \item In our formulation, we have to search for an optimal $\mbc$ (missing value splitting criteria) in addition to $\mbP$ and  $\mbq$, which are used for conventional non-missing splitting. A straightforward way to translate this into integer programming leads to a quadratic program. To make it linear, we add multiple intermediate variables, such as $w^{(1)}, w^{(2)}$ and $w^{\text{(nm)}}$.
\end{itemize}

\begin{program}[h]
\begin{align}
    &\text{minimize } \;\; \ell (\calD) + \lambda \cdot \ell_\text{fair} (\calD) \nonumber \\
    &q_v, \text{loss}(l), f_{l,0}, f_{l,1}, \ell_\text{fair}(\calD) \in \mathbb{R} \;\; \text{ for }  v \in \calV, l \in \calL \nonumber \\ 
    &w_{i,v}, w_{i,v}^{(1)}, w_{i,v}^{(2)}, w_{i,v}^{(\text{nm})},p_{v,j}, z_{i,l}, c_{v}, u_l \in \binset \;\; \text{ for } i \in [n], v \in \calV, j \in [d], l \in \calL \nonumber \\ 
    &\text{subject to:} \;\; \sum_{j \in [d]} p_{v,j} = 1,\sum_{l \in \mathcal{L}} z_{i,l} = 1 \label{eq:one_hot_const} \\ 
    & \qquad\qquad \;\;\; q_v - \sum_{j \in [d]} p_{v,j} (1-m_{i,j}) x_{i,j} \leq M  w_{i,v}^{(\text{nm})} - \epsilon(1- w_{i,v}^{(\text{nm})}) \label{eq:branch_const1} \\ 
    &\qquad\qquad \;\;\; q_v - \sum_{j \in [d]} p_{v,j} (1-m_{i,j}) x_{i,j} \geq -M (1- w_{i,v}^{(\text{nm})}) \label{eq:branch_const2} \\ 
    &\qquad\qquad \;\;\; w_{i,v}^{(1)} + 1 \geq (1 -  \sum_{j \in [d]} p_{v,j} m_{i,j}) + w_{i,v}^{(\text{nm})} \label{eq:branch_const3} \\ 
    &\qquad\qquad \;\;\; w_{i,v}^{(1)} \leq (1 -  \sum_{j \in [d]} p_{v,j} m_{i,j}), \;\; w_{i,v}^{(1)} \leq w_{i,v}^{(\text{nm})} \label{eq:branch_const4} \\ 
    &\qquad\qquad \;\;\; w_{i,v}^{(2)} + 1 \geq \sum_{j \in [d]} p_{v,j} m_{i,j} + c_v, \;\; w_{i,v}^{(2)} \leq \sum_{j \in [d]} p_{v,j} m_{i,j}, \;\; w_{i,v}^{(2)} \leq c_v \label{eq:branch_const5}\\ 
    &\qquad\qquad \;\;\; w_{i,v} \geq w_{i,v}^{(1)}, \;\; w_{i,v} \geq w_{i,v}^{(2)} \label{eq:branch_const6}\\ 
    &\qquad\qquad \;\;\; z_{i,l} \leq w_{i,v}, \;\; \forall l \in \mathcal{L}^L(v), \;\; z_{i,l} \leq 1- w_{i,v}, \;\; \forall l \in \mathcal{L}^R(v) \label{eq:branch_const7} \\ 
    &\qquad\qquad \;\;\;\sum_{i \in [n]} 2(y_i -1) z_{i,l} \leq M u_l - \epsilon (1-u_l) \label{eq:leaf_const1} \\
    &\qquad\qquad \;\;\; \sum_{i \in [n]} 2(y_i -1) z_{i,l} \geq M (1-u_l) - \epsilon (1-u_l) \label{eq:leaf_const2} \\
    &\qquad\qquad \;\;\; \text{loss}(l) \leq \sum_{i \in [n]} (1-y_i) z_{i,l}, \;\; \text{loss}(l) \leq \sum_{i \in [n]} y_i z_{i,l}  \label{eq:loss_const1} \\ 
    &\qquad\qquad \;\;\; \text{loss}(l) \geq \sum_{i \in [n]} (1-y_i) z_{i,l} - M (1-u_l), \;\; \text{loss}(l) \geq \sum_{i \in [n]} y_i z_{i,l} - M u_l \label{eq:loss_const2} \\
    &\qquad\qquad \;\;\; f_{l,0} \geq  \sum_{i \in [n]} (1-y_i) (1-s_i) z_{i,l} - M (1-u_l) \label{eq:fair_reg_const_1} \\
    &\qquad\qquad \;\;\; f_{l,0}  \leq \sum_{i \in [n]} (1-y_i) (1- s_i) z_{i,l} + M (1-u_l) + \epsilon \label{eq:fair_reg_const_2} \\ 
    &\qquad\qquad \;\;\; f_{l,0} \geq  \sum_{i \in [n]} y_i (1- s_i) z_{i,l} - M u_l, \;\;\; f_{l,0} \leq \sum_{i \in [n]} y_i (1- s_i) z_{i,l} + M u_l + \epsilon \label{eq:fair_reg_const_3} \\
    &\qquad\qquad \;\;\; f_{l,1} \geq  \sum_{i \in [n]} (1-y_i) s_i z_{i,l} - M (1-u_l) \label{eq:fair_reg_const_4} \\
    &\qquad\qquad \;\;\; f_{l,1}  \leq \sum_{i \in [n]} (1-y_i) s_i z_{i,l} + M (1-u_l) + \epsilon \label{eq:fair_reg_const_5} \\ 
    &\qquad\qquad \;\;\; f_{l,1} \geq  \sum_{i \in [n]} y_i s_i z_{i,l} - M u_l, \;\;\; f_{l,1} \leq \sum_{i \in [n]} y_i s_i z_{i,l} + M u_l + \epsilon\label{eq:fair_reg_const_6}, \\
    &\qquad\qquad \;\;\; \ell_\text{fair} (\calD) \geq  \frac{\sum_{l \in \calL} f_{l,0}}{\sum_{i \in [n]} (1-S_i) } - \frac{\sum_{l \in \calL} f_{l,1}}{\sum_{i \in [n]} s_i } \label{eq:l_fair_1} \\ 
    &\qquad\qquad \;\;\; \ell_\text{fair} (\calD) \geq -\left(\frac{\sum_{l \in \calL} f_{l,0}}{\sum_{i \in [n]} (1-S_i) } - \frac{\sum_{l \in \calL} f_{l,1}}{\sum_{i \in [n]} S_i } \right) \label{eq:l_fair_2}  \\
    &\qquad\qquad \;\;\; (\forall v \in \calV, \forall j \in [d], \forall l \in \calL, \forall i \in [n], \text{ except line \eqref{eq:branch_const7}}) \nonumber 
\end{align}
\caption{MIP Formulation for Fair Decision Tree Training with MIA}\label{prog:fair_tree}
\end{program}

\section{Experiment Details}\label{app:experiment}
\subsection{Datasets}
We use three different datasets for evaluation: COMPAS, Adult, and High School Longitudinal Study (HSLS). While COMPAS and Adult are widely used datasets in the fair machine learning literature, we believe that this paper is the first work to study the fair ML aspect of the HSLS dataset. We first give brief introduction to the dataset and then illustrate how missing patterns in the real-world survey data can have disparate missing patterns. 

\paragraph{Description of HSLS Dataset.} 
This dataset consists of 23,000+ participants from 944 high schools who were followed from the 9th through 12th grade.  It includes surveys from students, parents, and teachers, student demographic information, school information, and students academic performance across several years. The goal is to predict student's 9th-grade math test performance from relevant variables collected prior to the test. The original dataset has thousands of features, but for our analysis, we only utilize a set of 11 features: 
\begin{table}[h]
    \centering
    \begin{tabular}{clc}
    \toprule
       \textsc{Variable}  & \textsc{Description} & \textsc{Type} \\
    \midrule
    \textfn{X1RACE} & \sccell{l}{ Student's race/ethnicity } & Categorical  \\ 
    \midrule
    \textfn{X1MTHID} & \sccell{l}{Student's mathematics identity} & Continuous \\ 
    \midrule
    \textfn{X1MTHUTI} & \sccell{l}{Student's mathematics utility } & Continuous \\ 
    \midrule
    \textfn{X1MTHEFF} & \sccell{l}{Student's mathematics self-efficacy} & Continuous \\ 
    \midrule
    \textfn{X1PAR2EDU} & \sccell{l}{Secondary caregiver's highest level of education } & Categorical  \\  
    \midrule
    \textfn{X1FAMINCOME} & \sccell{l}{Total family income } & Continuous \\  
    \midrule
    \textfn{X1P1RELATION} & \sccell{l}{Relationship between student and the primary caregiver } & Categorical \\ 
    \midrule
    \textfn{X1PAR1EMP} & \sccell{l}{Primary caregiver's employment status } & Categorical\\  
    \midrule
    \textfn{X1SCHOOLBEL} & \sccell{l}{Student's sense of school belonging } & Continuous\\ 
    \midrule
    \textfn{X1STU30OCC2} & \sccell{l}{Student desired occupation at age 30 } & Categorical \\ 
    \midrule
    \textfn{X1TXMSCR} & \sccell{l}{Student's mathematics standardized test score }  & Continuous \\ 
    \bottomrule
    \end{tabular}
    \caption{Description of features used in the HSLS dataset}
    \label{tab:hsls_vars}
\end{table}

We use \textfn{X1RACE} to generate a binary group attribute: White/Asian (WA) and under-represented minority (URM) that includes Black, Hispanic, Native American, and Pacific Islanders. We create a binary label from the continuous test score \textfn{X1TXMSCR}, to perform a binary classification on whether a student belongs to the top 50\% performers or the bottom 50\% performers. As we do not consider the case when the group attribute or the label are missing, we drop data points that are missing \textfn{X1RACE} or \textfn{X1TXMSCR}. We scale every variable to be between 0 and 1.

\paragraph{Illustration of disparate missing patterns in the HSLS dataset.}
We show a real-world example of disparate missing patterns in the HSLS dataset. Between male and female students, male students consistently had 1-2\% more missing values in all variables. In Table~\ref{table:hsls_stat}, we summarize  missing probabilities of some variables between different demographic groups: male vs. female and  WA vs. URM. Between WA students and URM students, URM students always had higher missing probabilities in all variables. However, the difference in missing probabilities varied widely depending on the variables. Within the student survey variables, \textit{X1MTHID} had only 3\% difference between WA and URM, and \textit{X1MTHEFF} had about 6\% difference. For parent survey variables, URM consistently had 6-8\% higher missing probabilities. However, for the questions related to secondary caregivers, URM had a significantly higher missing rate, e.g. for \textit{X1PAR2EDU}, the difference was more than 15\%. On the other hand, between genders, while the  questions on secondary caregivers have a higher missing probability in general, the difference between males and females was not significant. Without an in-depth analysis, it is not straightforward to detect how missing patterns will vary depending on which group attributes.

\begin{table*}[ht]
\small
\centering
\renewcommand{\arraystretch}{0.3}
\begin{tabular}{l l c c c c}
\toprule
& & \multicolumn{4}{c}{\textsc{\textsc{Missing Probabilities}} (\%)} \\
\cmidrule(lr){3-4} \cmidrule(lr){5-6}

\textsc{Variable} & \textsc{Description} & \textsc{Male} & \textsc{Female} & \textsc{WA} & \textsc{URM}\\

\midrule
\textfn{X1MTHID} & \sccell{l}{ Student's mathematics identity } & 10.6 {\scriptsize $\pm$ 0.3} & 9.3 {\scriptsize$\pm$ 0.3} & 4.8 {\scriptsize$\pm$ 0.2} & 7.8 {\scriptsize$\pm$ 0.3}\\

\midrule

\textfn{X1MTHEFF} & \sccell{l}{Student's mathematics self-efficacy } & 21.2 {\scriptsize$\pm$ 0.4} & 19.1 {\scriptsize$\pm$ 0.4} & 14.0 {\scriptsize$\pm$ 0.3} & 20.1 {\scriptsize$\pm$ 0.4}\\

\midrule

\textfn{S1APCALC} & \sccell{l}{9th grader plans to enroll in an \\ Advanced Placement (AP) calculus course} & 14.2 {\scriptsize$\pm$ 0.3} & 12.1 {\scriptsize$\pm$ 0.3} & 7.6 {\scriptsize$\pm$ 0.2} & 12.1 {\scriptsize$\pm$ 0.3}\\

\midrule

\textfn{X1PAR1EDU} & \sccell{l}{Primary caregiver's highest level of education} & 29.6 {\scriptsize$\pm$ 0.4} & 27.5 {\scriptsize$\pm$ 0.4} & 22.8 {\scriptsize$\pm$ 0.4} & 29.8 {\scriptsize$\pm$ 0.5}\\

\midrule

\textfn{X1PAR2EDU} & \sccell{l}{Secondary caregiver's highest level of education} & 44.5 {\scriptsize$\pm$ 0.4} & 43.4 {\scriptsize$\pm$ 0.5} & 35.8 {\scriptsize$\pm$ 0.4} & \textbf{51.0} {\scriptsize$\pm$ 0.5}\\


\bottomrule
\end{tabular}
\caption{\small{Data missing probabilities of different variables between demographic groups: male vs. female or white vs. under-represented minority (URM) in the HSLS dataset}}
\label{table:hsls_stat}
\end{table*}

\paragraph{Description of COMPAS and Adult datasets.} For COMPAS dataset, we use eight features: \textit{age\_cat\_25-45}, \textit{age\_cat\_Greater-than-45}, \textit{age\_cat\_Less-than-25}, \textit{race},	\textit{sex}, \textit{priors\_count} \textit{c\_charge\_degree}, \textit{two\_year\_recid}. We use \textit{race} as a group attribute (0: Black, 1: White) and \textit{two\_year\_recid} as a binary label (1 indicates recidivating within two years and 0 otherwise). We balance the dataset to have equal number of Black and White subjects to isolate the issue of data imbalance out of our analysis. After balancing, we had a total of 4206 data points. 

For Adult dataset, we use the following features: \textit{age,	workclass,	education,	education-num,	marital-status,	occupation,	relationship,	race, gender, capital-gain,	capital-loss, hours-per-week, native-country,} and	\textit{income}. We use \textit{gender} as a group attribute (0: female, 1:male) and \textit{income} as a label (0: income $\leq$ 50K, 1: $>$ 50K). For Adult, we balance the dataset both in terms of the group attribute and the label, as the original dataset is highly imbalanced, having substantially more data points with label 0. After balancing, we had a total of 7834 data points.

As these two datasets do not contain any missing values, we artificially created missing values. The missing statistics we created is summarized below:
\begin{table}[h]
    \centering
  \begin{tabular}{c c c c}
  \toprule
Dataset & Feature & $p_0^{\text{ms}}$ & $p_1^{\text{ms}}$ \\ 
\midrule
\multirow{3}{*}{Adult} & marital-status & 0.0 & 0.4 \\ 
& hours-per-week & 0.0 & 0.3 \\ 
& race & 0.2 & 0.2 \\ 
\midrule
\multirow{2}{*}{COMPAS} & priors\_count & 0.4 & 0.1 \\ 
& sex & 0.6 & 0.2 \\
\bottomrule
\end{tabular}
    \caption{Missing statistics for Adult and COMPAS datasets we generated for our experiments. }
    \label{tab:art_miss_stat}
\end{table}

As the goal of the paper is to examine the fairness issues and their remedy, we tested different combinations of features with varying $p_0$  and $p_1$, where $p_0 = \Pr(M=1|S=0)$ and   $p_1 = \Pr(M=1|S=1)$ for each missing variable. We varied $p_0$  and $p_1$ from 0.0 to 0.9, and chose the missing pattern that had considerable difference in FNR of FPR between the two groups defined by the group attribute.

\subsection{Hyperparameters}
For the Fair MIP Forest algorithm, there are four hyperparameters we can choose: tree depth ($D$), number of trees ($n_\text{tree}$), time limit for training a single tree ($t_\text{limit}$), and the batch size. We chose $D = 3$ for all experiments as we did not see much improvement in accuracy when going from $D=3$ to $D=4,5$. We have tried $n_\text{tree} = 10, 20, 30, 40$,  $t_\text{limit} = 60, 90, 120$ (s), batch size $=200, 400, 800$, and picked the ones with best performance. If the performance was similar, we chose parameters that have smaller computational cost (e.g. smaller $n_\text{tree}$). The chosen hyperparameters are summarized below: 
\begin{table}[h]
    \centering
    \begin{tabular}{ccccc}
    \toprule
     Dataset & $t_\text{limit}$ & $n_\text{tree}$ & batch\_size & $\lambda$ \\ \midrule 
     COMPAS & 60 & 30  & 200 & \{0.1, 0.5, 1.0\} \\ \midrule 
     Adult & 60  &  30 & 200 &  \{0.1, 0.14, 0.17, 0.5, 0.8, 2.0 \}\\ \midrule 
     HSLS & 60 & 30 &  400 & \{ 0.1, 1.0 ,3.0 \} \\ \bottomrule
    \end{tabular}
    \caption{Summary of hyperparameters used in Fair MIP Forest.}
    \label{tab:hyperparameter}
\end{table}

We varied $\lambda$ in Fair MIP Forest algorithm to moderate how much we regularize fairness metrics. We varied $\lambda$ from 0.01 to 20. The final lambda values plotted in Figure~\ref{fig:experiment} are given in Table~\ref{tab:hyperparameter}.
For Zafar~\citep{zafar2019fairness}, we varied  $\tau$ to get different points on the fairness-accuracy trade-off plot. The set of $\tau$ values we tried is: \{0.001, 0.01, 0.1, 1, 10, 100\}. For Agarwal~\citep{agarwal2018reductions}, we varied $\epsilon$ to achieve different fairness-accuracy trade-off points, and the set of values used is: \{0.001, 0.005, 0.01, 0.02, 0.05, 0.1\}. For all methods, we drop points under the convex curve and only keep the best performing points on the plot.

For Agarwal and Hardt, we train a decision tree  classifier with the same parameters as the baseline (i.e. decision trees with depth 3). For Zafar, we use logistic regression, as it requires a distance-based classifier. 

\subsection{Implementation Details}
Agarwal and Hardt are implemented with AIF360~\citep{bellamy2018ai}. For Zafar, we use the code in \url{https://github.com/mbilalzafar/fair-classification}.

\end{document}